%% file: main.tex
\documentclass[10pt,twocolumn,letterpaper]{article}

\usepackage{cvpr}              

\definecolor{cvprblue}{rgb}{0.21,0.49,0.74}

\usepackage{multirow}
\usepackage{multicol}
\usepackage{amssymb}

\usepackage{pifont} 

\newcommand{\xmark}{\ding{55}}

\definecolor{emerald}{RGB}{80, 200, 120}
\definecolor{coral}{RGB}{255, 127, 80}
\definecolor{teal}{RGB}{0, 128, 128}
\definecolor{goldenrod}{RGB}{218, 165, 32}

\definecolor{darkgreen}{RGB}{0,100,0}
\definecolor{darkred}{RGB}{139,0,0}

\usepackage[pagebackref,breaklinks,colorlinks,allcolors=cvprblue]{hyperref}

\usepackage[table]{xcolor}
\usepackage{tcolorbox}

\title{Text-Driven Reasoning Video Editing via Reinforcement Learning on Digital Twin Representations}

\author{Yiqing Shen, Chenjia Li, Mathias Unberath\\
Department of Computer Science, Johns Hopkins University\\
{\tt\small \{yshen92, unberath\}@jhu.edu}
}

\begin{document}
\maketitle
\input{sec/0_abstract}    
\input{sec/1_intro}
\input{sec/2_relatedwork}
\input{sec/3_method}
\input{sec/4_exp}

\input{sec/5_conclusion}

\newpage

{
    \small
    \bibliographystyle{ieeenat_fullname}
    \bibliography{main}
}


\end{document}

%% file: sec/0_abstract.tex
\begin{abstract}

Text-driven video editing enables users to modify video content only using text queries.
While existing methods can modify video content if explicit descriptions of editing targets with precise spatial locations and temporal boundaries are provided, these requirements become impractical when users attempt to conceptualize edits through implicit queries referencing semantic properties or object relationships.
We introduce reasoning video editing, a task where video editing models must interpret implicit queries through multi-hop reasoning to infer editing targets before executing modifications, and a first model attempting to solve this complex task, RIVER (Reasoning-based Implicit Video EditoR).
RIVER decouples reasoning from generation through digital twin representations of video content that preserve spatial relationships, temporal trajectories, and semantic attributes.
A large language model then processes this representation jointly with the implicit query, performing multi-hop reasoning to determine modifications, then outputs structured instructions that guide a diffusion-based editor to execute pixel-level changes.
RIVER training uses reinforcement learning with rewards that evaluate reasoning accuracy and generation quality.
Finally, we introduce RVEBenchmark, a benchmark of 100 videos with 519 implicit queries spanning three levels and categories of reasoning complexity specifically for reasoning video editing.
RIVER demonstrates best performance on the proposed RVEBenchmark and also achieves state-of-the-art performance on two additional video editing benchmarks (VegGIE and FiVE), where it surpasses six baseline methods.

\end{abstract}

%% file: sec/1_intro.tex
\section{Introduction}

\begin{figure}[!htbp]
\centering
\includegraphics[width=\linewidth]{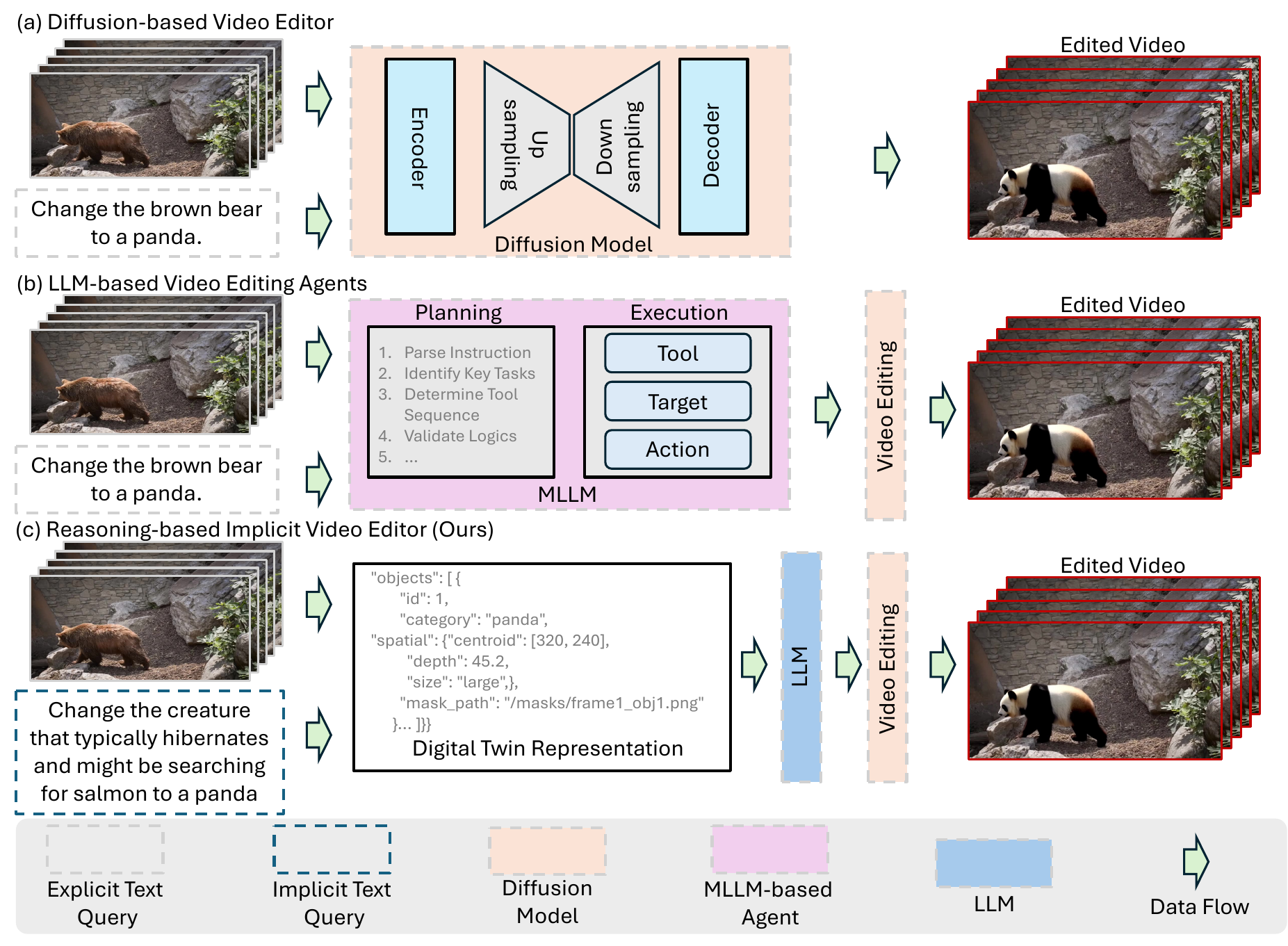}
\caption{Comparison of video editing paradigms. 
(a) Diffusion-based methods directly process explicit queries through conditional generation but struggle when editing targets require inference. 
(b) Agent-based approaches employ MLLMs for planning and execution but still depend on explicit specifications of what and where to edit. 
(c) Our proposed RIVER framework handles implicit queries by first constructing a digital twin representation where an LLM performs multi-hop reasoning to identify editing targets, then guides the diffusion model to execute modifications.
}
\label{fig:intro}
\end{figure}

Text-driven video editing with deep learning enables users, such as content creators or entertainment professionals, to modify videos using text queries, reducing the time and technical expertise required for traditional frame-by-frame editing~\cite{text2live2022,videdit2023}. 
Existing formulations of text-driven video editing~\cite{ccedit,videodirector,stablevideo} require explicit text queries that contain semantic descriptions of what to edit and when/where modifications should occur, typically relying on users to articulate both the visual content and its spatial/temporal context.
This formulation assumes that users can directly articulate visual attributes and exact positioning information for target objects or regions~\cite{via2024,expressedit2024},
, which can be challenging as users often struggle to precisely specify spatial-temporal details through text alone~\cite{expressedit2024,videodirector}.
Moreover, this requirement also becomes impractical for complex editing scenarios, where edits often involve multi-hop reasoning to infer the editing targets and their relationships with other elements~\cite{implicitqa2025,visa2024}.
To bridge this gap, we propose text-driven \textbf{reasoning video editing}, a task formulation in which video editing models must first reason about implicit queries to infer editing targets before executing the requested modifications.

Existing text-driven video editing methods can be grouped into three categories based on their approach to handling text queries. 
First, diffusion-based video editing methods~\cite{ccedit,dreamix} perform editing through conditional generation, where diffusion models synthesize modified content guided by both original video and text queries~\cite{contextualizeddiffusion2024,pix2video}. 
These methods assume that text queries provide explicit descriptions of editing targets with clear spatial and temporal specifications~\cite{videodirector}. 
When presented with complex editing tasks requiring spatial/temporal reasoning, these existing diffusion models can fail by editing incorrect objects, missing intended targets, or applying modifications to incorrect temporal segments~\cite{implicitqa2025,veggie2025,videodirector}. 
Second, video understanding models based on large language models (LLMs), such as VideoLLaMA~\cite{videollama2023} and Video-ChatGPT~\cite{videochatgpt2023} can interpret implicit queries and perform multi-hop reasoning on video content for tasks like question answering (QA) and captioning~\cite{videollama2-2024,mvbench2023}. 
However, these LLMs lack the generative abilities to modify video content based on their reasoning outputs~\cite{videomamba2024,internvideo2-2024}. 
They can understand what needs to be edited, but they cannot execute the edits.
Consequently, recent LLM-based video editing agents~\cite{lave2024,lstoryboard2025} attempt to bridge this gap by using LLM for high-level planning and task decomposition, then delegating to specialized editing tools such as diffusion models for execution.
For instance, LAVE~\cite{lave2024} uses LLMs to plan editing workflows and generate action sequences, while L-Storyboard~\cite{lstoryboard2025} converts video content into language representations for LLM processing.
Yet, these agents still require users to provide explicit specifications about editing targets and therefore cannot reason about implicit queries to identify targets before editing.
This reveals a gap between models that can reason about video content and models that can edit it, where no existing approach can interpret implicit queries and execute the corresponding edits, as depicted in Fig.~\ref{fig:intro}.

To address this gap, we propose a reasoning video editing framework (\textbf{R}easoning-based \textbf{I}mplicit \textbf{V}ideo \textbf{E}dito\textbf{R}, \textbf{RIVER}) that decouples reasoning from generation through an intermediate digital twin representation. 
Specifically, we construct a digital twin representation of the input video that preserves detailed scene information in a structured format. 
Unlike token-based video encoding in LLM, the digital twin representation maintains explicit spatial relationships between objects, temporal trajectories, and scene attributes that are important for reasoning about implicit queries. 
Afterwards, the digital twin representation and implicit editing query are processed by an LLM, which performs multi-hop reasoning to determine which objects or regions require modification and what transformations should be applied.
The LLM outputs structured editing instructions still in digital twin representation format that specify targets and operations. 
Finally, these instructions guide a diffusion-based video editor to execute the actual pixel-level modifications.
We train this architecture using reinforcement learning rather than supervised fine-tuning, which allows the model to learn from reward signals that evaluate both reasoning accuracy and generation quality.

Our contributions are three-fold. 
First, we introduce reasoning video editing, where models must interpret implicit text queries through multi-hop reasoning to identify editing targets before executing modifications. 
Second, we propose a framework that decouples reasoning from generation through digital twin representations.
Specifically, LLMs process these representations to infer editing targets and instructions, and reinforcement learning trains diffusion models and LLMs jointly to execute the edits while optimizing both reasoning accuracy and generation quality. 
Third, we introduce RVEBenchmark, a manually annotated benchmark comprising 100 videos with 519 implicit editing queries that span three levels of progressive reasoning complexity and three distinct reasoning categories (semantic, spatial, and temporal), where each query requires reasoning to determine editing targets before modification.

%% file: sec/2_relatedwork.tex
\section{Related Works}

\begin{figure*}[t!]
\centering
\includegraphics[width=\linewidth]{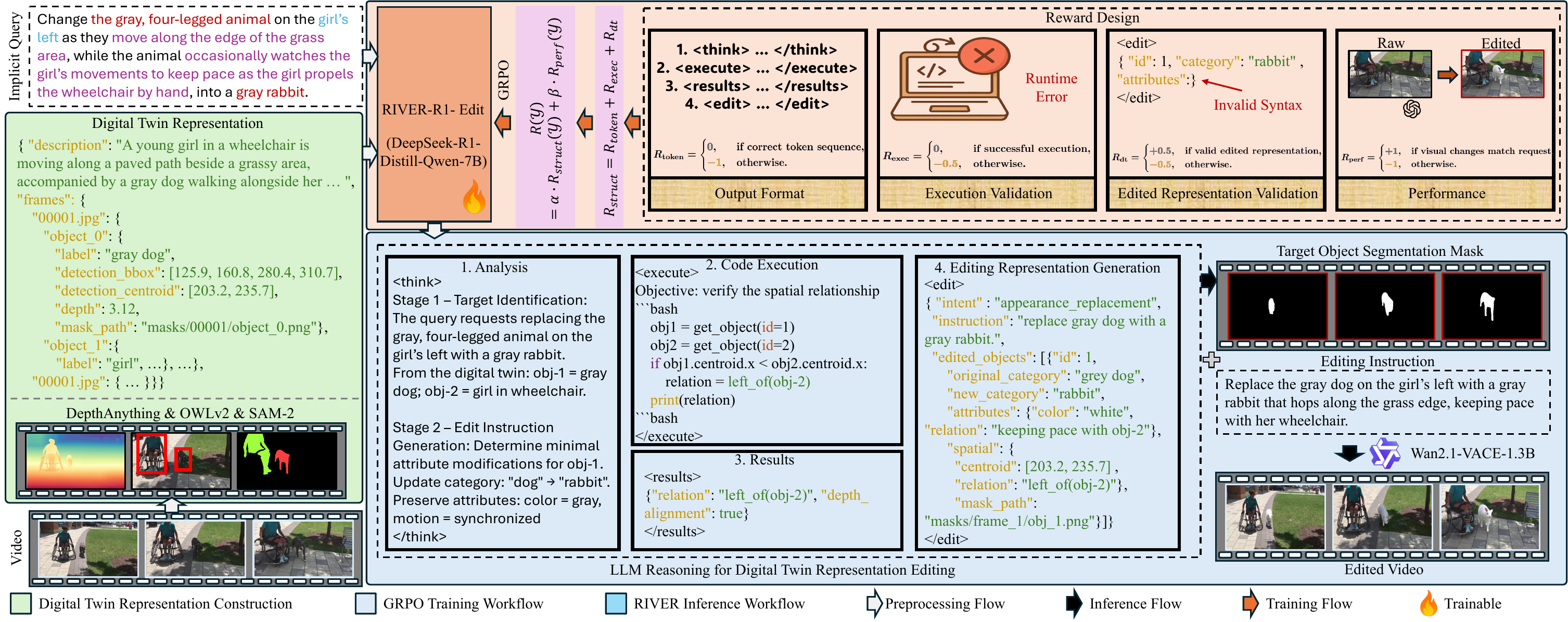}
\caption{Overview of RIVER framework. 
Green: Digital twin representation construction extracts structured scene information from video frames using specialist vision models.
Orange: GRPO training workflow updates the LLM through reward signals that validate output format correctness ($R_{\text{token}}$), code executability ($R_{\text{exec}}$), edited representation validity ($R_{\text{dt}}$), and generation performance ($R_{\text{perf}}$). 
Blue: Inference workflow processes implicit queries through four stages.
LLM performs reasoning analysis (\texttt{<think>}), generates executable code for spatial-temporal verification (\texttt{<execute>}), evaluates computation results (\texttt{<results>}), and produces edited digital twin representations (\texttt{<edit>}) to guide the diffusion model to synthesize the final edited video.
}
\label{fig:method}
\end{figure*}

\paragraph{Text-Driven Video Editing.}
Diffusion models formulate video editing as a conditional generation problem.
For example, CC-Edit~\cite{ccedit}, Dreamix~\cite{dreamix}, StableVideo~\cite{stablevideo}, Pix2Video~\cite{pix2video}, InstructVid2Vid~\cite{instructvid2vid}, and EffiVED~\cite{effived} leverage cross-attention to inject text embeddings from queries into the denoising process.
However, they require the queries to explicitly specify editing targets with clear spatial locations and temporal boundaries, limiting their applicability when users express intentions through indirect references or semantic descriptions.
Instruction-based editing approaches such as SmartEdit~\cite{smartedit} extend editing capabilities by combining multimodal LLM (MLLM) with diffusion models to interpret complex queries that involve reasoning.
However, this work operates only on static images and does not generalize to videos.
To address the need for processing complex editing requests for videos, agent-based methods introduce planning capabilities that leverage LLMs to decompose editing queries into structured workflows.
For example, LAVE~\cite{lave2024} uses LLMs to plan editing workflows and generate action sequences for specialized tools such as video retrieval, clip sequencing, and trimming based on user objectives.
Similarly, L-Storyboard~\cite{lstoryboard2025} converts video content into language representations for LLM processing to decompose editing tasks into structured operations and generate storyboards for narrative construction.
Yet, current agents still require users to explicitly state which objects or regions need modification and where those changes should occur.

\paragraph{Video Reasoning with MLLMs.}
MLLMs~\cite{videochatgpt2023} perform reasoning over videos by combining visual encoders with LLM for question-answering and captioning tasks~\cite{momentor, mmbenchwideo}.
Specifically, they process semantic relationships and temporal dynamics in videos through language to answer questions that require reasoning.
More recent work demonstrates that MLLMs can perform implicit query reasoning to identify editing targets without explicit spatial coordinates or temporal boundaries~\cite{survey}.
For example, VISA~\cite{visa2024} introduces reasoning video object segmentation (ReasonVOS), where LLMs leverage world knowledge and video context to interpret complex queries and reason about spatial and temporal relationships to localize target objects.
However, these MLLMs lack the video generative abilities needed to modify video content based on their reasoning outputs.
Furthermore, token-based video encoding in MLLM compresses visual information into fixed-length embeddings through spatial pooling or projection, discarding fine-grained spatial relationships and object boundaries~\cite{position}.
Therefore, it prevents precise localization of target regions and boundaries, hindering the integration of reasoning with generation for video editing.

\paragraph{Digital Twin Representations.}
Digital twin representations are outcome-driven intermediate representation that preserve semantic, spatial, and temporal information from raw data such as videos~\cite{position}.
Unlike digital twins that maintain bidirectional synchronization with physical entities, digital twin representations serve as static structured abstractions created from observations~\cite{position}.
Furthermore, unlike token representations that compress visual information through direct vectorization, digital twin representations maintain explicit relationships between entities and their interactions, making them more suitable for downstream tasks requiring multi-hop reasoning~\cite{position,jit}.
Recent work demonstrates that digital twin representations can bridge perception and reasoning for video analysis tasks such as reasoning segmentation~\cite{jit,survey}.
For example, just-in-time digital twins construct digital twin representations on-demand using specialist vision models, where LLM subsequently performs reasoning on top of them to determine the object for segmentation~\cite{jit}.
Digital twin representations have been applied to temporally-constrained reasoning segmentation~\cite{shen2025temporally}, operating room workflow analysis~\cite{shen2025operating}, and automated benchmark construction for visual reasoning tasks~\cite{shen2025rvtbench}.
Motivated by these advantages, our work adopts digital twin representations as an intermediate layer between reasoning and generation for video editing by allowing LLMs to perform multi-hop reasoning on explicit object relationships before guiding diffusion-based editing.

%% file: sec/3_method.tex
\section{Methods}

\begin{figure*}[t!]
\centering
\includegraphics[width=\linewidth]{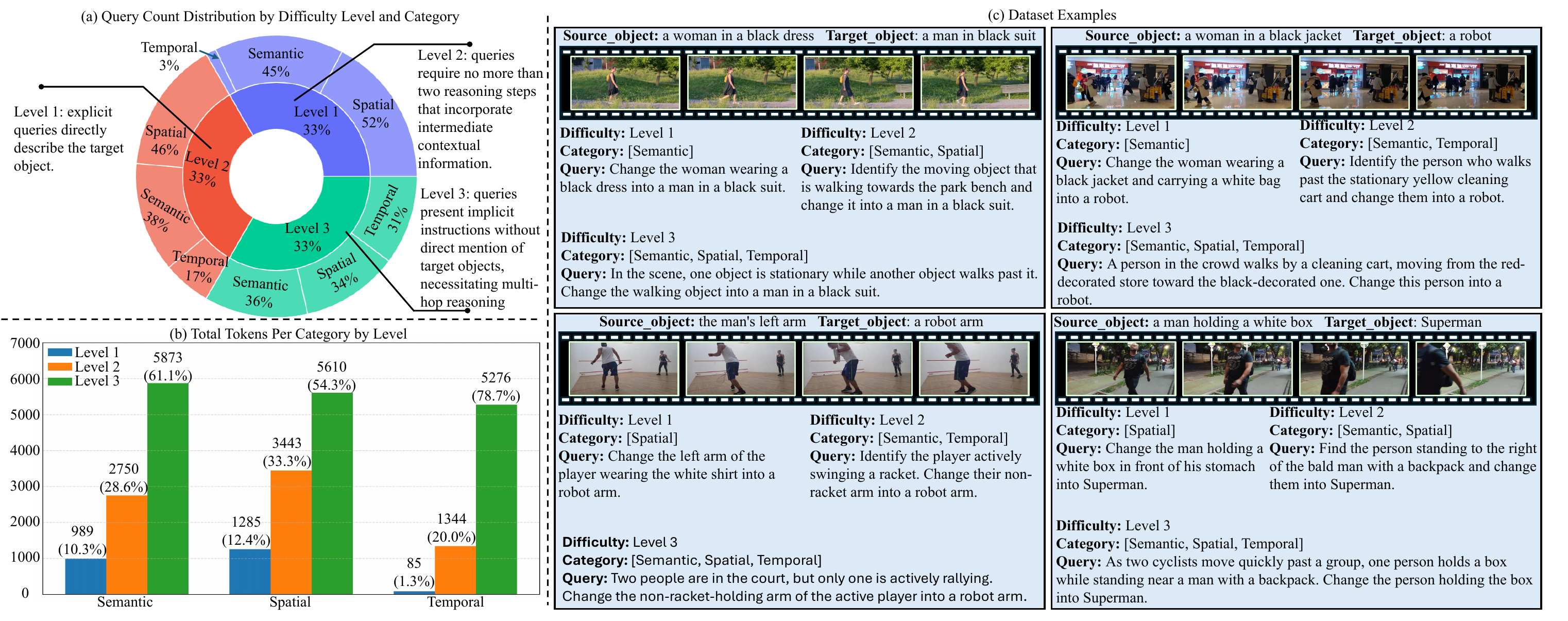}
\caption{Overview of RVEBenchmark dataset structure and examples.
(a) Distribution of 519 queries across three difficulty levels, where each level combines different reasoning categories. 
%
%
(b) Token distribution reveals that Level 3 queries contain the most tokens (61.1\% for semantic, 54.3\% for spatial, 78.7\% for temporal), reflecting their increased complexity. 
(c) Four representative examples demonstrate progressive reasoning difficulty.}
\label{fig:dataset}
\end{figure*}
 
\subsection{RIVER Model}
\paragraph{Overview.}
We propose a reasoning video editing framework, called \textbf{RIVER} (\textbf{R}easoning-based \textbf{I}mplicit \textbf{V}ideo \textbf{E}dito\textbf{R}) , which decouples reasoning from video editing through intermediate digital twin representations, as shown in Fig.~\ref{fig:method}.
Given an input video $\mathcal{V} = \{I_t\}_{t=1}^T$ consisting of $T$ frames and an implicit editing query $q$, our goal is to produce an edited video $\mathcal{V}'$ that satisfies the user's intent expressed in $q$. 
The framework operates in three stages.
First, we construct a digital twin representation $\mathcal{D}$~\cite{position} from the input video using vision foundation models~\cite{sam1,sam2,depthanything,qwenvl}, which encode the semantic attributes, spatial relationships, and temporal dynamics of all objects in a structured text format.
This digital twin representation, combined with the implicit editing query, is then fed into an LLM that performs multi-hop reasoning where the editing targets are identified before the generation of the edited digital twin representation $\mathcal{D}'$  specifying desired modifications.
In other words, the LLM directly modifies the digital twin representation rather than the video itself, allowing edits to be reasoned and planned at a semantic level before execution.
Subsequently, this edited digital twin representation guides the diffusion-based video editor to synthesize the final output. 
We train this architecture end-to-end using reinforcement learning via GRPO~\cite{grpo}, since acquiring paired training data containing both reasoning chains and corresponding video edits is impractical for supervised fine-tuning. 
The training optimizes the reward containing two components, namely (1) structured rewards that enforce correct reasoning chains and output formats from the LLM, and (2) performance rewards that evaluate generation quality from the diffusion model through LLM-as-a-judge~\cite{zheng2023judging}.

\paragraph{Digital Twin Representation.}
The digital twin representation transforms raw video frames into structured descriptions that preserve fine-grained object information while remaining accessible for LLM-based reasoning.
Specifically, for a video $\mathcal{V} = \{I_t\}_{t=1}^T$ containing $T$ frames, we construct a digital twin representation $\mathcal{D} = \{D^{(1)}, D^{(2)}, \cdots , D^{(T)}\}$ where each frame representation $D^{(t)}$ decomposes the scene into object-level components.
Each frame representation contains a set of detected object instances, formalized as $D^{(t)} = \{(i, c_i^{(t)}, a_i^{(t)}, m_i^{(t)}, s_i^{(t)})\}_{i=1}^{N^{(t)}}$, where $N^{(t)}$ denotes the number of instances detected in frame $t$.
Each instance tuple includes an identifier $i$ that maintains correspondence between frames, a semantic category $c_i^{(t)}$ assigned by object detection models, attribute descriptions $a_i^{(t)}$ that capture visual properties such as color and texture, a mask path $m_i^{(t)}$ that references instance segmentation, and spatial properties $s_i^{(t)} = (x, y, d, \text{size})$ that encode centroid coordinates, depth values, and object dimensions.
We employ vision foundation models to extract these components, where SAM-2~\cite{sam2} generates instance-level segmentation masks while tracking objects between frames, DepthAnything~\cite{depthanything} estimates per-pixel depth maps that we sample at object centroids, OWLv2~\cite{owlv2} assigns semantic categories to detected instances, and Qwen2.5-VL~\cite{qwenvl} generates natural language descriptions of object attributes.
The complete digital twin representation $\mathcal{D}$ is serialized in the JSON format, which enables LLMs to process scene information directly through their language understanding without requiring visual token compression.

\paragraph{LLM Reasoning for Digital Twin Representation Editing.}
Given the digital twin representation $\mathcal{D}$ of video $\mathcal{V}$ and the implicit query $q$, we employ an LLM to perform multi-hop reasoning that identifies editing targets and determines required editing operations.
The LLM generates structured output following a specific rollout sequence, where it begins by analyzing the query within \texttt{<think>} and \texttt{</think>} tokens. 
Through this chain-of-thought reasoning, the LLM identifies target objects $\mathcal{O}_{\text{edit}} = \{i_1, i_2, ..., i_K\}$ that satisfy the editing query $q$.
When spatial or temporal computations are required, LLM generates executable Python code within \texttt{<execute>} and \texttt{</execute>} tokens that operates directly on the digital twin representation.
The execution results appear within \texttt{<results>} and \texttt{</results>} tokens, enabling iterative refinement where initial findings inform subsequent reasoning steps.
Once the editing targets are determined, the LLM generates an edited digital twin representation $\mathcal{D}'$ within \texttt{<edit>} and \texttt{</edit>} tokens by modifying relevant object attributes, spatial properties or instance relationships in the JSON structure.
Such editing on the digital twin representation allows the specification of transformations such as changing object colors, adjusting spatial positions, or removing instances in an explicit text manner without directly manipulating pixel values.
For each modified object, the edited representation $\mathcal{D}'$ preserves the structural format of the original while updating the specific fields that correspond to the requested edits.

\paragraph{Digital Twin Representation Guided Video Editing.}
The final stage of RIVER employs a latent diffusion model to synthesize edited video $\mathcal{V}'$ from original video $\mathcal{V}$ and edited digital twin representation $\mathcal{D}'$.
Unlike standard text-to-video diffusion models that are conditioned solely on natural language queries, we condition the diffusion process on the digital twin representation $\mathcal{D}'$, which provides precise object-level specifications compared to free-form text descriptions.
To convert the structured JSON format of $\mathcal{D}'$ into conditioning signals for the diffusion model, we first parse the edited digital twin to extract two types of information: textual descriptions and spatial guidance. 
For textual conditioning, we serialize the modified object attributes $(c_i^{(t)}, a_i^{(t)})$ from $\mathcal{D}'$ into natural language descriptions following the template such as ``\textit{In frame $t$, object $i$ with category $c_i^{(t)}$ has attributes $a_i^{(t)}$.}" 
These descriptions are encoded by the text encoder of the diffusion model, producing text embeddings.
For spatial conditioning, we load the segmentation masks referenced by $m_i^{(t)}$ in $\mathcal{D}'$ to identify the pixel regions that require modification specified by LLM. 
These masks, combined with the spatial properties $s_i^{(t)}$, form spatially aligned conditioning maps that are concatenated with the noisy latent representation in each denoising step.
This spatial guidance directs the diffusion model to apply edits precisely to the specified object regions while preserving the appearance of unmodified areas. 
Consequently, the edited video output $\mathcal{V}'$ maintains the temporal dynamics and visual quality of the original while incorporating the modifications specified in $\mathcal{D}'$.

\begin{table*}[t]
\centering
\caption{Quantitative evaluation of reasoning video editing methods on RVEBench.  
Each metric is assessed across three levels of reasoning complexity in percentage (\%). 
Higher scores indicate better performance for all metrics. 
RIVER demonstrates competitive performance across all reasoning levels.
}
\label{tab:evaluation_metrics}
\resizebox{\textwidth}{!}{%
\begin{tabular}{@{}l*{21}{c}@{}}
\toprule
\multirow{3}{*}{Method} 
& \multicolumn{3}{c}{CLIP-Text ($\uparrow$)} 
& \multicolumn{3}{c}{CLIP-F ($\uparrow$)} 
& \multicolumn{3}{c}{MUSIQ ($\uparrow$)} 
& \multicolumn{3}{c}{SSIM ($\uparrow$)} 
& \multicolumn{3}{c}{PSNR ($\uparrow$)}
& \multicolumn{3}{c}{GroundingDINO ($\uparrow$)} 
& \multicolumn{3}{c}{LLM-as-a-Judge ($\uparrow$)} \\
\cmidrule(lr){2-4} \cmidrule(lr){5-7} \cmidrule(lr){8-10} 
\cmidrule(lr){11-13} \cmidrule(lr){14-16} \cmidrule(lr){17-19} \cmidrule(lr){20-22}
& L1 & L2 & L3 & L1 & L2 & L3 & L1 & L2 & L3 
& L1 & L2 & L3 & L1 & L2 & L3 & L1 & L2 & L3 & L1 & L2 & L3 \\
\midrule
InstructDiff~\cite{geng2024instructdiffusion}     
& 19.57 & 18.23 & 17.21 & 86.80 & 86.40 & 86.17 & 42.22 & 42.02 & 41.62 
& 60.34 & 60.13 & 59.20 & 16.02 & 15.85 & 15.49 & 14.73 & 10.08 & 9.38 
& 44.33 & 39.89 & 36.79 \\
InstructV2V~\cite{instructvid2vid}      
& 22.22 & 21.43 & 19.78 & 91.80 & 91.44 & 90.59 & 42.65 & 44.43 & 44.14  
& 41.44 & 41.41 & 40.72 & 10.89 & 11.07 & 10.96 & 11.48 & 7.50 & 5.13 
& 38.86 & 35.95 & 34.70 \\
FlowDirector~\cite{li2025flowdirector}     
& 19.56 & 18.70 & 17.15 & 93.90 & 94.06 & 94.10 & 47.20 & 47.29 & 47.33 
& 61.90 & 62.37 & 62.19 & 19.63 & 20.11 & 20.01 & 8.73 & 8.11 & 5.56 
& 35.52 & 30.75 & 29.96 \\
AnyV2V~\cite{ku2024anyv2v}           
& 17.05 & 16.29 & 15.54 & 93.71 & 93.85 & 93.61 & 46.73 & 46.56 & 46.64 
& 49.79 & 49.12 & 48.28 & 15.14 & 15.03 & 14.92 & 13.72 & 12.38 & 9.94 
& 20.01 & 18.47 & 17.68 \\
InstructPix2Pix~\cite{brooks2023instructpix2pix}  
& 18.49 & 17.45 & 16.73 & 92.18 & 92.73 & 93.06 & 45.40 & 44.90 & 45.37  
& 61.75 & 61.94 & 61.66 & 15.38 & 15.47 & 15.08 & 6.19 & 6.32 & 9.68 
& 34.53 & 30.76 & 29.49 \\
GPT-Image-1      
& 26.55 & 26.10 & 25.81 & 87.65 & 88.68 & 88.96 & 51.09 & 51.06 & 51.56 
& 44.47 & 42.11 & 40.76 & 13.40 & 12.94 & 12.37 & 25.93 & 19.23 & 24.00 
& 68.12 & 66.55 & 65.18 \\
\midrule
RIVER (Ours) & \textbf{27.85} & \textbf{26.93} & \textbf{26.17} & \textbf{97.45} & \textbf{96.32} & \textbf{96.04} & \textbf{53.13} & \textbf{52.83} & \textbf{52.97} & \textbf{77.08} & \textbf{74.12} & \textbf{72.95} & \textbf{20.06} & \textbf{19.81} & \textbf{19.17} & \textbf{47.14} & \textbf{43.79} & \textbf{39.28} & \textbf{82.10} & \textbf{79.18} & \textbf{73.89} \\
\bottomrule
\end{tabular}%
}
\end{table*}

\begin{table*}[t!]
\centering
\caption{Quantitative evaluation of video editing methods on VegGIE dataset~\cite{veggie2025}.
%
We report mean and standard deviation across seven metrics in percentages (\%).
Our RIVER demonstrates competitive performance across all evaluation metrics.
}
\label{tab:evaluation_metrics_veggie}
\resizebox{0.95\textwidth}{!}{%
\begin{tabular}{@{}lccccccc@{}}
\toprule
Method 
& CLIP-Text ($\uparrow$) 
& CLIP-F ($\uparrow$) 
& MUSIQ ($\uparrow$) 
& SSIM ($\uparrow$) 
& PSNR ($\uparrow$)
& GroundingDINO ($\uparrow$)
& LLM-as-a-Judge ($\uparrow$) \\
\midrule
InstructDiff~\cite{geng2024instructdiffusion}     
& 21.20{\tiny$\pm3.12$} & 90.92{\tiny$\pm2.32$}\ & 51.35{\tiny$\pm3.40$} & 44.23{\tiny$\pm13.57$} & 13.47{\tiny$\pm0.48$} & 25.00{\tiny$\pm16.76$} & 40.20{\tiny$\pm11.67$} \\
InstructV2V~\cite{instructvid2vid}      
& 20.35{\tiny$\pm2.79$} & 92.90{\tiny$\pm2.03$} & 31.17{\tiny$\pm4.17$} & 39.32{\tiny$\pm12.34$} & 11.41{\tiny$\pm0.68$} & 43.75{\tiny$\pm19.15$} & 37.33{\tiny$\pm10.42$} \\
FlowDirector~\cite{li2025flowdirector}     
& 20.67{\tiny$\pm2.58$} & 96.33{\tiny$\pm0.83$} & 53.77{\tiny$\pm3.29$} & 63.99{\tiny$\pm16.40$} & 16.08{\tiny$\pm1.60$} & 20.00{\tiny$\pm13.25$} & 42.27{\tiny$\pm10.11$} \\
AnyV2V~\cite{ku2024anyv2v}           
& 19.18{\tiny$\pm3.40$} & 95.28{\tiny$\pm0.76$} & 56.36{\tiny$\pm3.01$} & 67.11{\tiny$\pm11.93$} & 16.02{\tiny$\pm1.50$} & 31.25{\tiny$\pm15.62$} & 29.55{\tiny$\pm11.60$} \\
InstructPix2Pix~\cite{brooks2023instructpix2pix}  
& 22.20{\tiny$\pm3.12$} & 91.36{\tiny$\pm3.80$} & 53.77{\tiny$\pm3.29$} & 43.00{\tiny$\pm12.49$} & 14.18{\tiny$\pm0.78$} & 37.50{\tiny$\pm23.27$} & 37.06{\tiny$\pm10.93$} \\
GPT-Image-1 & 23.86{\tiny$\pm3.39$} & 90.50{\tiny$\pm3.12$} & 63.38{\tiny$\pm2.59$} & 39.41{\tiny$\pm14.94$} & 11.77{\tiny$\pm0.82$} & 53.33{\tiny$\pm17.17$} & 79.19{\tiny$\pm8.99$}\\
\midrule
RIVER (Ours) 
& \textbf{27.43{\tiny$\pm2.14$}} & \textbf{97.41{\tiny$\pm0.60$}} & \textbf{68.43{\tiny$\pm4.27$}} & \textbf{71.06{\tiny$\pm6.46$}} & \textbf{17.81{\tiny$\pm1.52$}} & \textbf{75.00{\tiny$\pm9.46$}} & \textbf{92.95{\tiny$\pm1.83$}} \\
\bottomrule
\end{tabular}%
}
\end{table*}

\subsection{Reinforcement Learning}

\paragraph{Reward Functions.}
We design a reward function $R$ to train the RIVER model, \textit{i}.\textit{e}., LLM and diffusion model jointly through GRPO~\cite{grpo}, where the total reward combines structured rewards $R_{\text{struct}}$ that validate the output format of LLM and performance rewards $R_{\text{perf}}$ that evaluate editing performance of the diffusion model.
Formally, the reward is defined as $R = \alpha R_{\text{struct}} + \beta R_{\text{perf}}$, where $\alpha$ and $\beta$ are balancing coefficients.
The structured reward $R_{\text{struct}}$ comprises three components that verify the LLM output format, namely $R_{\text{struct}}=R_{\text{token}}+R_{\text{exec}}+R_{\text{dt}}$.
First, the token format reward $R_{\text{token}}$ assigns $0$ if all required token pairs appear in the correct sequence, including \texttt{<think>}, \texttt{<execute>}, \texttt{<results>}, and \texttt{<edit>} with their corresponding closing tokens, otherwise $-1$.
Second, the code executability reward $R_{\text{exec}}$ validates that Python operations within \texttt{<execute>} tokens can be executed without syntax or runtime errors, assigning $0$ for successful execution and $-0.5$ for failures.
Third, the digital twin representation format reward $R_{\text{dt}}$ verifies that the edited representation $\mathcal{D}'$ within \texttt{<edit>} tokens maintains valid JSON structure and preserves the schema of the original representation $\mathcal{D}$, assigning $+0.5$ if valid and $-0.5$ otherwise.
Third, the digital twin representation format reward $R_{\text{dt}}$ verifies that the edited representation $\mathcal{D}'$ within \texttt{<edit>} tokens maintains valid JSON structure and preserves the schema of the original representation $\mathcal{D}$, assigning $+0.5$ if valid and $-0.5$ otherwise. 
Specifically, $\mathcal{D}'$ is valid if and only if the JSON string must be parseable without syntax errors, all required fields from the schema must be present for each object instance, and field types must match their specifications in the original representation. 
The performance reward $R_{\text{perf}}$ evaluates the quality of the final edited video $\mathcal{V}'$ through LLM-as-a-judge~\cite{zheng2023judging}, where we prompt a separate MLLM to assess whether the edits in $\mathcal{V}'$ align with the intent expressed in the implicit query $q$, assigning $+1$ for correct edits and $-1$ for incorrect ones.
Formally, the MLLM judge receives both the original video $\mathcal{V}$ and the edited video $\mathcal{V}'$ as visual input along with the editing query $q$, and is asked to determine whether the observed visual changes match the requested modifications while preserving unrelated scene elements.

\paragraph{Training.}
The training data consists of tuples $(V, q)$, where $V$ represents the input video, $q$ denotes the implicit editing query. 
Following previous work~\cite{segzero}, we do not include any reasoning data for cold start training, since collecting paired data with explicit reasoning chains and video edits is impractical.
During training, we freeze the denoising network parameters in the diffusion model while updating both the text encoder within the diffusion model and the LLM parameters through LoRA~\cite{lora}. 
We train RIVER using GRPO~\cite{grpo}, where the LLM generates multiple rollout sequences for each training sample, producing diverse reasoning chains and edited digital twin representations.
The reward function $R$ evaluates each rollout by computing structured rewards that validate output formats and performance rewards that assess editing quality. 
GRPO then updates the LLM policy by maximizing expected rewards while maintaining proximity to a reference policy through KL divergence regularization.

\subsection{Benchmark Dataset}
We construct \textbf{RVEBenchmark}, a manually annotated benchmark designed to evaluate reasoning video editing capabilities, as shown in Fig.~\ref{fig:dataset}. 
The benchmark builds upon videos from DAVIS~\cite{davis} and SA-V~\cite{sam2} datasets, comprising 100 videos paired with 519 implicit editing queries, where each query requires reasoning to identify editing targets before executing modifications.
RVEBench adopts the annotation protocol established by JiTBench~\cite{jit}, organizing queries by both levels of complexity of reasoning and categories of reasoning. 
We define three progressive levels of reasoning difficulty~\cite{shen2025rvtbench}. 
Level 1 queries involve single-step reasoning, where the target object is described directly, requiring models to identify the object based on explicit attributes. 
Level 2 queries require no more than two reasoning steps that incorporate intermediate contextual information, where models must connect multiple semantic or spatial relationships to determine editing targets.
Level 3 queries present implicit instructions without direct mention of target objects, necessitating multi-hop reasoning that operate on abstract cues such as functional relationships, contextual associations, or implied references. 
Beyond complexity levels, each query belongs to one of three reasoning categories, namely semantic reasoning requires understanding object attributes and conceptual relationships; spatial reasoning involves interpreting geometric arrangements and positional relationships between objects; temporal reasoning focuses on motion patterns, event sequences, and changes across frames.
Consequently, each benchmark sample contains the source video sequence at original resolution, an implicit editing query, the designated reasoning complexity level, the corresponding reasoning category, and binary segmentation masks that delineate target objects to be edited. 

%% file: sec/4_exp.tex
\section{Experiments}

\begin{table*}[t]
\centering
\caption{Quantitative evaluation of video editing methods on FiVE dataset~\cite{li2025five}.
%
}
\label{tab:evaluation_metrics_five}
\resizebox{\textwidth}{!}{%
\begin{tabular}{@{}lccccccc@{}}
\toprule
Method 
& CLIP-Text ($\uparrow$) 
& CLIP-F ($\uparrow$) 
& MUSIQ ($\uparrow$) 
& SSIM ($\uparrow$) 
& PSNR ($\uparrow$)
& GroundingDINO ($\uparrow$)
& LLM-as-a-Judge ($\uparrow$) \\
\midrule
InstructDiff~\cite{geng2024instructdiffusion}     
& 24.81{\tiny$\pm0.28$} & 88.03{\tiny$\pm0.36$} & 49.15{\tiny$\pm1.62$} & 54.53{\tiny$\pm0.26$} & 19.28{\tiny$\pm0.38$} & 17.67{\tiny$\pm4.07$}  & 54.83{\tiny$\pm7.11$} \\
InstructV2V~\cite{instructvid2vid}      
& 25.31{\tiny$\pm0.25$} & 91.84{\tiny$\pm0.28$} & 47.68{\tiny$\pm1.71$} & 50.12{\tiny$\pm0.38$} & 13.79{\tiny$\pm0.20$} & 14.67{\tiny$\pm3.96$} & 59.85{\tiny$\pm10.50$} \\
FlowDirector~\cite{li2025flowdirector}     
& 20.50{\tiny$\pm0.44$} & 96.91{\tiny$\pm0.11$} & 50.23{\tiny$\pm1.74$} & 61.39{\tiny$\pm0.13$} & 17.33{\tiny$\pm0.38$} & 19.59{\tiny$\pm7.50$} & 37.20{\tiny$\pm8.16$} \\
AnyV2V~\cite{ku2024anyv2v}           
& 24.73{\tiny$\pm0.36$} & 96.98{\tiny$\pm0.12$} & 55.64{\tiny$\pm1.89$} & 47.28{\tiny$\pm0.79$} & 18.06{\tiny$\pm0.46$} & 20.00{\tiny$\pm5.26$} & 54.85{\tiny$\pm8.76$}\\
InstructPix2Pix~\cite{brooks2023instructpix2pix}  
& 23.22{\tiny$\pm0.27$} & 92.26{\tiny$\pm0.25$} & 46.59{\tiny$\pm1.67$} & 58.27{\tiny$\pm0.14$} & 18.10{\tiny$\pm0.42$} & 22.81{\tiny$\pm4.45$} & 41.85{\tiny$\pm12.34$}\ \\
GPT-Image-1 & 28.42{\tiny$\pm1.21$} & 91.42{\tiny$\pm2.94$} & 57.55{\tiny$\pm3.28$} & 52.44{\tiny$\pm3.05$} & 15.05{\tiny$\pm1.49$} & 22.22{\tiny$\pm13.39$} & 68.81{\tiny$\pm7.06$}\\
\midrule
RIVER (Ours) 
& \textbf{33.01{\tiny$\pm0.63$}} & \textbf{97.57{\tiny$\pm0.17$}} & \textbf{61.42{\tiny$\pm2.09$}}  & \textbf{66.89{\tiny$\pm0.57$}} & \textbf{21.49{\tiny$\pm0.55$}} & \textbf{37.78{\tiny$\pm7.22$}} & \textbf{78.00{\tiny$\pm13.66$}} \\
\bottomrule
\end{tabular}%
}
\end{table*}

\begin{table*}[t]
\centering
\caption{Ablation study of RIVER components on RVEBenchmark across reasoning complexity levels.
We evaluate the contribution of each component.
Checkmarks (\checkmark) indicate the component is included, while crosses (\xmark) indicate removal.
Results demonstrate that all components contribute to overall performance, with digital twin representation and reinforcement learning rewards being particularly important for complex reasoning tasks.
}
\label{tab:ablation_study}
\resizebox{\textwidth}{!}{%
\begin{tabular}{@{}cccc*{21}{c}@{}}
\toprule
\multirow{3}{*}{\shortstack{Digital\\Twin}} 
& \multirow{3}{*}{\shortstack{LLM\\Reasoning}}
& \multirow{3}{*}{\shortstack{Structured\\Rewards}}
& \multirow{3}{*}{\shortstack{Performance\\Rewards}}
& \multicolumn{3}{c}{CLIP-Text ($\uparrow$)} 
& \multicolumn{3}{c}{CLIP-F ($\uparrow$)} 
& \multicolumn{3}{c}{MUSIQ ($\uparrow$)} 
& \multicolumn{3}{c}{SSIM ($\uparrow$)} 
& \multicolumn{3}{c}{PSNR ($\uparrow$)}
& \multicolumn{3}{c}{GroundingDINO ($\uparrow$)} 
& \multicolumn{3}{c}{LLM-Judge ($\uparrow$)} \\
\cmidrule(lr){5-7} \cmidrule(lr){8-10} \cmidrule(lr){11-13} 
\cmidrule(lr){14-16} \cmidrule(lr){17-19} \cmidrule(lr){20-22} \cmidrule(lr){23-25}
& & & 
& L1 & L2 & L3 & L1 & L2 & L3 & L1 & L2 & L3 
& L1 & L2 & L3 & L1 & L2 & L3 & L1 & L2 & L3 & L1 & L2 & L3 \\
\midrule
\xmark & \checkmark & \checkmark & \checkmark    
& 24.05 & 23.73 & 23.13 & 97.12 & 96.15 & 96.99 & 50.05 & 49.75 & 50.63 
& 64.37 & 64.86 & 65.63 & 19.86 & 19.59 & 19.47 & 44.00 & 41.15 & 38.74 
& 56.69 & 53.03 & 52.77 \\
\checkmark & \xmark & \checkmark & \checkmark      
& 25.49 & 25.13 & 24.77 & 97.09 & 96.28 & 96.83 & 51.48 & 51.24 & 51.31  
& 65.25 & 65.17 & 66.19 & 19.99 & 19.04 & 19.42 & 45.00 & 42.36 & 39.25 
& 74.44 & 71.18 & 66.54 \\
\checkmark & \checkmark & \xmark & \checkmark  
& 24.01 & 23.25 & 23.11 & 97.14 & 96.96 & 96.87 & 52.90 & 51.91 & 52.07 
& 65.48 & 65.88 & 64.69 & 19.94 & 19.58 & 19.45  & 46.15 & 43.09 & 40.44 
& 71.40 & 67.39 & 63.20 \\
\checkmark & \checkmark & \checkmark & \xmark     
& 24.34 & 23.44 & 22.15 & 96.94 & 95.89 & 95.61 & 52.21 & 50.99 & 51.59 
& 65.39 & 66.03 & 64.89 & 19.73 & 19.75 & 19.27 & 43.48 & 41.17 & 38.00 
& 63.27 & 58.61 & 54.03 \\
\midrule
\checkmark & \checkmark & \checkmark & \checkmark 
& 27.85 & 26.93 & 26.17 & 97.45 & 96.32 & 96.04 & 53.13 & 52.83 & 52.97 & 67.08 & 66.12 & 66.95 & 20.06 & 19.81 & 19.57 & 47.14 & 43.79 & 39.28 & 82.10 & 79.18 & 73.89 \\

\bottomrule
\end{tabular}
}
\end{table*}

\begin{figure}[htbp!]
\centering
\includegraphics[width=\linewidth]{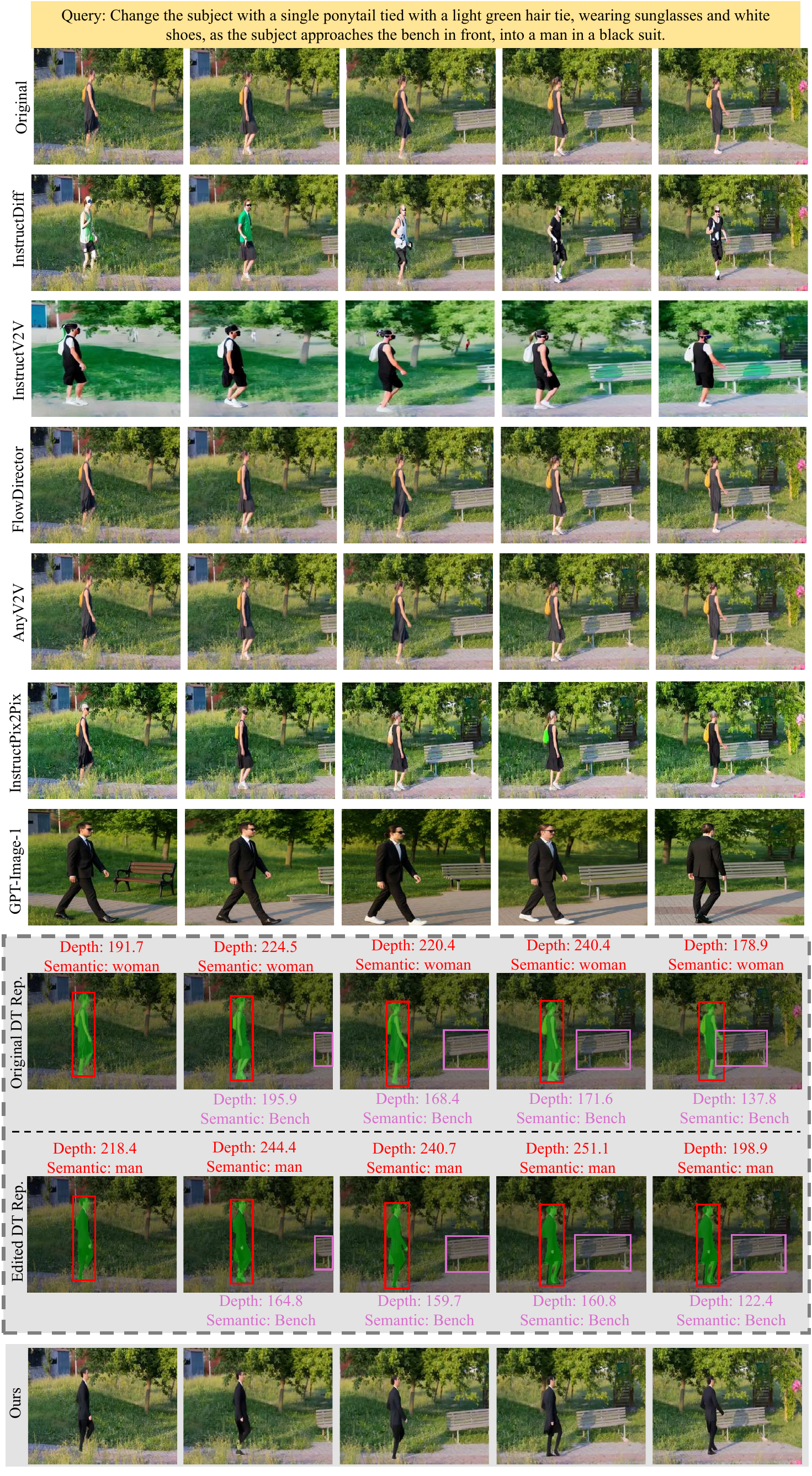}
\caption{Qualitative comparison of RIVER against six baseline methods on RVEBenchmark. 
The implicit query requires multi-hop reasoning to identify the target subject based on visual attributes, spatial relationships, and semantic understanding. 
%
%
The bottom three rows shown the original digital twin representation, the edited digital twin representation from LLM, and RIVER's final edited video frames. 
%
}
\label{fig:result}
\end{figure}

\paragraph{Implementation Details.}
We implement RIVER using DeepSeek-R1-Distill-Qwen-7B~\cite{r1} as the LLM for reasoning and Wan2.1-VACE-1.3B~\cite{wan2025} as the diffusion model for video editing on PyTorch version 2.4.0.
We train RIVER using GRPO~\cite{grpo} with a batch size of 8, a learning rate of 5e-7, and a total epoch of 10 using 510 training samples from RVTBench~\cite{shen2025rvtbench}.
We set the LoRA~\cite{lora} rank at 8, and the balancing coefficients $\alpha$ and $\beta$ in the reward function at 1.
We conduct all experiments on 4 A100 NVIDIA GPUs with 80 GB memory each.
All videos are resized to $832\times480$ resolution during both training and inference for consistency.

\paragraph{Compared Methods and Metrics.}
We compare RIVER with six baseline methods spanning image and video editing approaches. 
For video editing approaches based on diffusion models, we include three recent methods: InstructV2V~\cite{instructvid2vid}, which performs end-to-end diffusion-based editing; FlowDirector~\cite{li2025flowdirector}, an inversion-free framework that operates directly in data space using ODE-guided spatiotemporal manifolds for precise localized edits; and AnyV2V~\cite{ku2024anyv2v}, a tuning-free paradigm that leverages arbitrary image editing models and image-to-video generation through temporal feature injection. 
We additionally compare against two image editing baselines adapted for video through frame-by-frame processing: InstructDiff~\cite{geng2024instructdiffusion}, a generalist diffusion model that interprets human instructions to manipulate images; and InstructPix2Pix~\cite{brooks2023instructpix2pix}, which learns to follow image editing instructions via conditional diffusion model. 
We also compare against GPT-Image-1, an MLLM-based approach originally designed for instruction-guided image editing, where we adapt it to video by processing frames sequentially as individual images
For evaluation, we employ seven metrics.
Specifically, CLIP-Text and CLIP-F~\cite{hessel2021clipscore,wu2025fame,veggie2025} measure semantic and visual alignment between edited content and queries.
MUSIQ~\cite{ke2021musiq}, SSIM~\cite{wang2004image}, and PSNR~\cite{huynh2008scope} assess video quality, structural similarity, and signal fidelity between edited and original videos.
GroundingDINO~\cite{liu2023grounding} provides open-set object localization accuracy by detecting whether target objects specified in implict editing queries are correctly localized in edited outputs.
Finally, LLM-as-a-Judge~\cite{zheng2023judging} with GPT-5 assesses alignment of the output images with implicit queries.
All metrics are reported as percentages (\%) for consistent comparison.

\paragraph{Benchmark Datasets.}
We evaluate RIVER on three datasets, namely our proposed RVEBenchmark, VegGIE~\cite{veggie2025}, and FiVE~\cite{li2025five}.
RVEBenchmark contains 100 videos with 519 implicit queries in three categories and levels of reasoning, designed explicitly for reasoning video editing tasks.
VegGIE~\cite{veggie2025} offers 132 video query pairs that span 8 diverse editing skills, including concept addition, removal, object change, and stylization, paired with dedicated metrics for each skill.
FiVE provides 100 videos with 420 object-level query pairs across six fine-grained editing types, testing precise object-level modifications while maintaining temporal consistency.
Together, these benchmarks allow for a thorough assessment of both implicit reasoning abilities and explicit editing capabilities.

\paragraph{Evaluation on RVEBenchmark.}
Table~\ref{tab:evaluation_metrics} presents quantitative results on RVEBenchmark. 
RIVER achieves the best performance across all metrics. 
%
%
For LLM-as-a-Judge evaluation, RIVER obtains 82.10\%, 79.18\%, and 73.89\%, demonstrating that edits align with user intent substantially better than GPT-Image-1's 68.12\%, 66.55\%, and 65.18\%. 
Fig.~\ref{fig:result} illustrates a representative example where RIVER correctly identifies and transforms the target object, while compared methods either fail to locate the correct target or produce temporally inconsistent edits. 
These results confirm that decoupling reasoning from generation through digital twin representations enables RIVER to handle implicit queries requiring complex multi-step inference.

\paragraph{Evaluation on VegGIE dataset.}
Table~\ref{tab:evaluation_metrics_veggie} shows the results on VegGIE.
RIVER achieves the best performance across all metrics, with the CLIP-Text score of 27.43\% surpassing the second-best baseline by 15.0\%, demonstrating our method's capacity to interpret and execute varied editing operations accurately.
The GroundingDINO score of 75.00\% represents a 40.6\% improvement over the next best approach, indicating that our LLM-based reasoning effectively localizes correct editing targets across different manipulation types.
The LLM-as-a-Judge score of 92.95\% confirms that RIVER produces edits that align closely with user intent across different editing categories, outperforming GPT-Image-1's score of 79.19\%.
%
%
These results validate that the proposed RIVER generalizes well across diverse editing scenarios.

\paragraph{Evaluation on FiVE dataset.}
As shown in Table~\ref{tab:evaluation_metrics_five}, RIVER achieves the best performance in all metrics.
The dataset's emphasis on precise object-level modifications highlights our digital twin representation's advantages, as evidenced by RIVER's GroundingDINO score of 37.78, which exceeds the next best baseline by 65.6\%, indicating better capability in accurately localizing and modifying specific objects without affecting surrounding regions. 
Our method maintains high temporal coherence with a CLIP-F score of 97.57, demonstrating that the our frame-to-frame object tracking effectively preserves temporal consistency even when executing fine-grained edits. 
The high PSNR value of 21.49 further confirms that RIVER preserves visual fidelity during object-level transformations, outperforming diffusion-based approaches that often struggle with reconstruction quality when making localized changes. 
%

\paragraph{Ablation Study.}
We conduct ablation study, as shown in Table~\ref{tab:ablation_study}, which shows the importance of each component. 
For example, removing the digital twin representation results in degraded performance across all metrics, which demonstrates that preserving explicit spatial relationships through structured scene representations enables better target localization compared to token-based encoding. 
%

%% file: sec/5_conclusion.tex
\section{Conclusion}
We introduce reasoning video editing, where models interpret implicit text queries to infer editing targets before executing modifications. 
RIVER addresses this task by separating reasoning from generation through digital twin representations. 
This intermediate layer allows LLMs to perform multi-hop reasoning over structured scene information, then guide diffusion models to execute pixel-level changes.
Our work shows that digital twin representations~\cite{position} can effectively mediate between reasoning and generation. 
This opens up directions for reasoning-driven content creation, where text descriptions of abstract relationships can guide precise visual transformations. 
Future work can explore how these methods scale to longer videos and interactive editing workflows where users iteratively refine their intent through dialog.